# Computational Register Analysis and Synthesis

**Shlomo Engelson Argamon**
Department of Computer Science
Illinois Institute of Technology
Chicago, IL 60616
`argamon@iit.edu`

**Abstract**
The study of register in computational language research has historically been divided into *register analysis*, seeking to determine the registerial character of a text or corpus, and *register synthesis*, seeking to generate a text in a desired register. This article surveys the different approaches to these disparate tasks. Register synthesis has tended to use more theoretically articulated notions of register and genre than analysis work, which often seeks to categorize on the basis of intuitive and somewhat incoherent notions of prelabeled "text types". I argue that a integration of computational register analysis and synthesis will benefit register studies as a whole, by enabling a new large-scale research program in register studies. It will enable comprehensive global mapping of functional language varieties in multiple languages, including the relationships between them. Furthermore, computational methods together with high coverage systematically collected and analyzed data will thus enable rigorous empirical validation and refinement of different theories of register, which will have also  implications for our understanding of linguistic variation in general.

## 1. Introduction

Register, broadly construed, refers to variation in language usage within different functional contexts. This article surveys research on register specifically within *computational language research*. By this term I mean to include work in (a) computational linguistics, which seeks to understand language by reference to implementing computational models, (b) natural language processing, which seeks to develop automated methods for understanding, extracting information from, and generating natural language texts, and (c) corpus linguistics, which uses computational methods applied to large bodies of text to study linguistic structure. This article will consider how register variation is addressed computationally in all of these areas, not always distinguishing among them.

Historically, most computational language research has sidestepped the question of register, either by working with a "representative corpus" of text, however defined, or by limiting attention to one or a few specific text types. While there is widespread acknowledgement of the obvious differences between spoken and written text, and between formal and informal registers (say, published articles versus social media), until relatively recently, the computational study of linguistic register was a niche area and received little attention in computational work on language overall.

My aim in this article is to make sense of the plethora of diverse approaches to register from a computational perspective and to suggest how future research might most productively be pursued.

Within computational language research, work on register can be divided into two broad categories, based on the research goals (cf. Figure 1). The first is *register analysis*, studying how to computationally classify texts according to register, as well as how to analyze register characteristics computationally. The second is *register synthesis*, developing techniques to generate texts in a given register, either starting from a formal semantic representation, or by "translating" texts from one register to another.

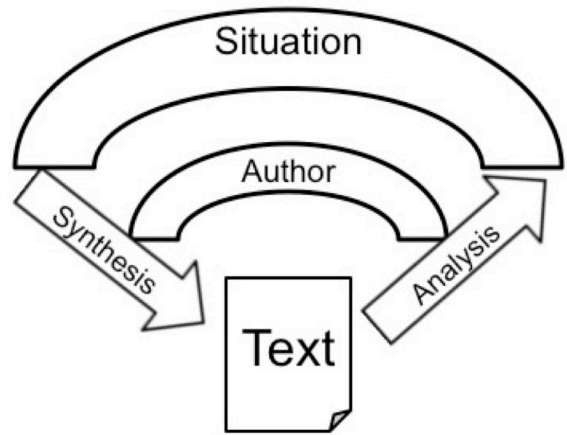

*Figure 1.* Register analysis and synthesis.

Given a corpus of texts labeled according to different registers, computationally classifying new texts according to register is a well-defined task. It does, however, require a definition of register sufficiently precise that human annotators can label texts accordingly with high inter-rater reliability, which is not always easy to achieve. Register classification can comprise a stand-alone task (Stamatatos, Fakotakis, & Kokkinakis 2000, Biber & Conrad 2001, Argamon, Koppel, Fine, & Shimoni 2003, Finn & Kushmerick 2006, Santini 2006, Herring & Paolillo 2006, Abbasi & Chen 2007, Dong, Watters, Duffy, & Shepherd 2008, Sharoff, Wu, & Markert 2010) or may be used to derive insights into larger questions related to linguistic variation (e.g., Atkinson 1992, Argamon, Dodick, & Chase 2008, Eisenstein, Smith, & Xing 2011, Teich, Degaetano-Ortlieb, Kermes, & Lapshinova-Koltunski 2013, Clarke & Grieve 2017). Register labels, either manually or automatically assigned, can also be used to control for register in research on other text analysis methods (e.g., Carroll et al. 1999, Giesbrecht & Evert 2009, Sharoff et al. 2010); differences in register between training and testing data often affect outcomes for NLP tasks such as part-of-speech tagging, parsing, or information extraction. Register categories can also be used to improve information retrieval and search, by serving as a constraint on what documents are retrieved (Karlgren 1999, Morato, Llorens, Génova, & Moreiro 2003, Amasyalı & Diri 2006, Freund, Clarke, & Toms 2006).

In text generation, explicit parameterization of register (in terms of communication goals, domain of discourse, level of formality, etc.) can be used as constraints on the process of generating natural language text, to achieve more natural and useable texts (DiMarco & Foster 1997, Kan & McKeown 2002, Power, Scott, & Bouayad-Agha 2003, Reiter & Williams 2010). This requires, of course, an elaborated theory of how register parameters are realized in text by either restrictive or preferential constraints on word choice, syntactic structures, and so on. In research on register translation by analogy to machine translation between different languages, statistical models of the relationship between linguistic characteristics of different registers can be applied to a source text to generate a semantically equivalent text in a different register (Ficler & Goldberg 2017, Jhamtani, Gangal, Hovy, & Nyberg 2017, Fu, Tan, Peng, Zhao, & Yan 2017).

Related work on computer-mediated communication (CMC) has also been significant in the study of register, since the rise of new communication modes and media due to computational communication systems has enabled clearer and more precisely targeted research on the evolution and emergence of new registers (Nowson, Oberlander, & Gill 2005, Herring & Paolillo 2006, Crystal 2011). Much of this work falls outside our scope of computational language research, as it uses traditional (non-computational) methods of textual analysis. However, corpus linguistic methods are commonly used, and some CMC registers (e.g., blogs and Twitter) are frequent testbeds for NLP research.

## 2. Defining Register

Unfortunately, there is a fair bit of terminological fuzziness and inconsistency in the computational literature dealing with register. For many natural language processing researchers, "register" is often used as a rough synonym of the vague concept "text type", by which is meant a group of texts sharing some more-or-less identifiable set linguistic characteristics, without any clear theoretical motivation. Indeed, in much of this literature, the term "register" can be used interchangeably with "genre" or "text type", and little attention is given to the structure of register or its categories. For the most part, this sort of informal approach is taken when studies control experiments for register/genre/text type, often using labels pre-assigned by corpus developers.

Indeed, most register classification research, which seeks to determine which of a discrete set of register categories a text belongs to, usually proceeds atheoretically. Register categories are sometimes based on reliable, if simple, external features of the text (spoken vs. written, for example), sometimes on other theoretical grounds such as the community of discourse or purpose of the text, and sometimes based on clusters of shared linguistic characteristics. Register categories considered in such studies tend to be chosen for convenience, e.g., based on what categories are provided by a particular corpus such as the British National Corpus (2007), or for the specific needs of a particular study, such as comparison of, e.g., peer-reviewed scientific papers vs. science popularizations. Little attention is usually placed on the functional relationship between linguistic features of register categories and the needs of their typical situations of use, though at times analysis results are interpreted in light of such functional considerations (Argamon et al. 2008, Teich et al. 2013).

For our purposes, we adopt the perspective of Biber and Conrad (2009: 6) who define register as "a [language] variety associated with a particular situation of use (including particular communicative purposes)." A register is described by that situational context and the linguistic features typical of the register, along with a description of how those features function specifically within that particular context of language use. That is, the linguistic features describing a register are not arbitrary, but form a complex that is useful for particular communicative purposes in a particular context. We note that register is thus primarily defined by the situations in which it is used and only secondarily by the linguistic features by which it is recognized. This perspective is not universally recognized within computational language research, however.

Where computational language research has historically grappled seriously with the concept of register, generally one of two theoretical perspectives is taken. Both views accord broadly with Biber and Conrad, in that they take register variation as essentially functional in nature, deriving from the fact that different communicative contexts require different linguistic resources to be brought to bear, resulting in (statistically) different language varieties. They differ, however, in how they structure register descriptions, how they differentially foreground either the situational context or the internal linguistic features of the text, and in how they formulate the relationship of register to other aspects of linguistic variation such as genre and style.

The main theoretically-motivated approach taken within register analysis is the basis of the multidimensional approach to register developed by Biber (e.g., 1991), which grew out of the definition above. This approach seeks to encode register character by a set of natural dimensions of variation for a mass of linguistic variables. Each dimension defines a spectrum of one aspect of register variation, such as formal-informal, narrative-informational, etc. Large numbers of linguistic features covary along each dimension, such that the set of dimensions explains much of the variation in linguistic form between registers and individual texts. With focus on the internal linguistic consistency of registers, the method remains ecumenical with respect to what external constraints (social situation, domain, function, historical context, etc.) are relevant to register. Two fundamental notions in this view are (a) that register variation is (or can be) continuous, varying along multiple fundamental dimensions of variation, and that (b) registers come at different levels of generality, with more general registers not being strictly comparable to more specific registers.

The theoretical view which predominates in work on register synthesis, on the other hand, is that of Systemic Functional Linguistics (SFL), which elaborates on the situational context by considering register to comprise distinct language varieties that serve differentiated social and communicative functions (Halliday, McIntosh, & Strevens 1968). This notion of register was elaborated by Martin (1992) to be determined by the contextual variables of *field* (the type and domain of social discourse), *tenor* (the relationship between the speaker and audience, and their relevant social roles), and *mode* (parameters of textual organization, such as the communication channel and discourse goals). Each of these constrains the typical linguistic features that will be used in a text, based on functional requirements such as politeness, clarity, precision, available information bandwidth, and so forth. This articulation is particularly congenial for work on text generation systems, where values for field, tenor, and mode can be set manually, and researchers work on translating those into methods for determining the manner in which internal semantic representations are to be realized in text, for example choosing between complex or simple syntactic structures (Bateman 1997). Discourse-level generation work (e.g., Hovy 1991, Moore & Paris 1993) using Rhetorical Structure Theory (Mann & Thompson, 1988) also relies on this concept of register.

## 2. Computational Register Analysis

### 2.1. Aims of Research in Computational Register Analysis

Register analysis seeks to determine or characterize the register(s) of a text or of a collection of texts. A key task is to determine the character of particular registers, and their relationships to each other, in terms of what distributions of linguistic features they are realized by. Effective methods for register analysis can also be used for numerous applications including improving text analysis accuracy by controlling for register (Carroll et al. 1999, Kakkonen & Sutinen 2008, Giesbrecht &

Evert 2009, Sharoff et al. 2010, Rehbein & Bildhauer 2017), information retrieval focused on particular registers and genres (Karlgren 1999, Crowston & Kwasnik 2003, Morato et al. 2003, Amasyalı & Diri 2006, Freund et al. 2006, Vidulin, Luštrek, & Gams 2007), and analyzing the evolution of linguistic norms in different discourse communities (Biber & Finegan 2001, Speelman, Gondelaers, & Geeraerts 2006, Argamon et al. 2008, Teich & Fankhauser 2011, Degaetano-Ortlieb et al. 2016).

There are two main approaches to register analysis: classification, which seeks to divide texts into distinct register categories, and multidimensional analysis, which seeks to find a set of continuous dimensions, each representing a spectrum of underlying register-relevant variation.

**2.1.1 Aims of Register Classification**

Traditionally, registers have usually been thought of as discrete categories of texts, such as spoken vs. written (at the most general level) or news stories vs. editorials (at a more specific level). This viewpoint has been adopted by much computational work on analysis of register, viewing it fundamentally as a classification or clustering problem of determining which texts fall into which register categories. As mentioned above, this work varies greatly in its theoretical approach to register, from essentially atheoretical approaches that treat register as a vague notion of "text type" to those that have a stronger theoretical treatment of register emerging either from context of creation/use of the text (external) or clusters of cooccurring linguistic features (internal). In much of this work, little distinction is made between the terms "genre" and "register"; since the computational methods for classifying texts based on genre or register differ little, we will discuss both types of work interchangeably here. We do note that this lack of theoretical rigor in this work is problematic and discuss this issue further below.

The main goal of work on computational register analysis, viewing registers as discrete categories, has been to develop methods for accurately classifying texts into their proper register categories. As such, the register analysis problem is formulated as a text classification problem (Sebastiani 2002). As such, the task is generally split into two subtasks: extracting useful features from the text and building a classification model using those features. In register classification, as in most text classification research, the focus has been on feature extraction, as generic classification machine learning methods are used. The question then is to determine a set of textual features that (a) are useful for a given register classification task and (b) can be accurately extracted using extant natural language processing methods.

Little work, if any, has examined this problem specifically from the standpoint of register, as opposed to other stylistic questions such as genre and authorship – researchers have sought "stylistic features" of texts that are correlated with different styles, contrasted with "topical features" that correlate with different topics or domains of discourse, which are typically used for information retrieval and related tasks. A wide variety of such features have been proposed (Stamatatos et al. 2000, Finn & Kushmerick 2006, Argamon & Koppel 2010), including relative frequencies of function words, part-of-speech n-grams, character n-grams, syntactic constructs, and systemic-functional categories, as well as type/token ratios, word and sentence length, and other textual statistics. These are discussed in more detail below.

The classification methods described above are all *supervised* methods, in that the training data are all labeled with the desired register categories. There is also some work on using *unsupervised* learning (i.e., clustering) for register analysis. These methods seek to find natural register categories in a corpus, to find the categories in a "bottom-up" way, rather than relying on manual annotation from a set of predefined register labels. Results are evaluated by comparing the sets of documents in each cluster to sets derived from manually annotated data. Since the number of clusters output is a parameter, care must be taken to compare results for different numbers of output clusters, evaluating the clusters for plausibility compared to the manual annotations. Clustering, as opposed to supervised categorization, may be advantageous when seeking to characterize the different registers that may evolve with the emergence of new communications mediums such as web pages, blogs, or the like.

Gries, Newman, and Shaoul (2011) performed a systematic study of clustering over n-gram features in two corpora to study how well n-grams can be used as features to distinguish registers and sub-registers. Their analysis found considerable structure in the corpora, with clustering giving results that matched well to fine-grained sub-registers, with 3-grams (word triples) giving overall the best results. Other clustering studies include Santini's influential studies of the genre characteristics of web pages (Santini 2005, 2006, 2008). Her results showed clear differentiation of register/genre categories emerging in an evolving computer-mediated communication medium. Clustering can also be used as an adjunct to classification analysis, to validate stability of results, as used by Teich and Fankhauser (2010) in their study of register differences between articles in different scientific fields (this study is discussed in more detail below).

When successful classification for a particular register task is achieved, examination of the features most influential in that classification can lend some insight into the linguistic difference between the registers being studied. For example, Biber and Barbieri (2007) show that different university registers use different characteristic lexical bundles (recurrent sequences of words), which they theorize realize different discourse frames and stances required by the different registers.

In this vein, register classification has also been used as to address social science research questions, by determining if two bodies of text from different communities differ linguistically, and whether such differences (if found) support theories about relevant differences in how the communities conceptualize, communicate, or organize information. For example, studies of variation in written and spoken registers of various World Englishes (Szmrecsanyi 2009, Gries & Mukherjee 2010) have shown clear differences between varieties of English in fundamental linguistic character such as analyticity/syntheticity and point towards how such analysis can help elucidate evolutionary relationships between these varieties. Classification studies of scientific writing from different disciplines have given insight into how the scientific register evolves (Atkinson 1992) and how different fields construct and communicate knowledge differently (Argamon et al. 2008, Teich et al. 2013).

**2.1.2 Aims of Multidimensional Analysis**

Douglas Biber (1991) pioneered another view into the nature of register and its relationship with underlying linguistic features, on the one hand, and related notions such as genre and text type, on the other. The central idea is that registers (and genres) differ along various continuous dimensions of covariation in sets of linguistic features. While the positioning of different registers at different points on each dimension critically can be related to aspects of their different contexts-of-use, the multidimensional approach methodologically takes a text-internal view of register, considering as primary the linguistic features characterizing different registers, irrespective of external situational variables.

Initial research in this paradigm focused on extracting and validating what the primary such dimensions are and correlating them with existing register and genre categories. The main idea is, viewing texts as points in a high-dimensional vector space (whose axes are linguistic variables), to use statistical techniques such as principal components analysis (PCA) (Jolliffe 2011) to determine the main directions of stylistic variation in that space. Subsequently, linguistic analysis of the main variables participating in each dimension, along with consideration of what kinds of texts have what scores in each dimension, can give a meaningful characterization of the spectrum that each dimension covers, such as "involved—informational" (factor 1) and "narrative—non-narrative" (factor 2). Different register categories reflect choices of regions in each of these dimensions; a multidimensional analysis of a text gives a fine-grained picture of the overall register/genre character of the text.

It is worth noting that the features considered in most applications of this approach are local lexico-grammatical features – lexical items, parts-of-speech, choices of syntactic structure – and do not include features of the overall discourse structure of the text. This focus accords with the view (Biber et al. 2007) that views register as realized through lexico-grammatical variation, as distinguished from genre, which is largely realized via variation in discourse structure. It might be useful to probe this distinction computationally, by using existing methods to extract aspects of rhetorical and discourse structure (Marcu 2000) and use them in multidimensional analysis to see what macrostructures might (or might not) covary with lexico-grammatical features between registers.

Biber's original factors as well as his methodology have been applied to register and genre analysis in a number of domains after his original study. The factors in that study have emerged from analysis of diverse corpora, corroborating the general nature of that work (Lee 2001). Work applying multidimensional analysis to different corpora has been used to study the evolution of and variation in scientific writing (Atkinson 1992, 1998, Conrad 1996), the register and genre structure of blogs (Grieve et al. 2010), how textual cohesion is differentially realized in register (Louwerse & Graesser 2004), how interpersonal stance markers relate to register distinctions (Pavalanathan, Fitzpatrick, Kiesling, & Eisenstein 2017), and more.

Briefly, we note the stream of computational work that restricts the definition of "register" to the single dimension of formality, defining register as the level of formality of the language. Formality distinguishes well between spoken and written registers, as well as between personal registers (such as letters and diaries) and published registers (such as fiction and non-fiction essays and books). Formality measures such as Heylighen and Dewaele's (1999) F-score can be used to estimate the formality of a text for corpus linguistic studies (Brooke, Wang, & Hirst 2010, Mosquera & Moreda 2012, Sheika & Inkpen 2012). This perspective can be viewed as a choosing a single Biberian dimension for analysis and use.

*2.2 Methods of Register Analysis Research*

Register analysis, whether in the categorical or the multidimensional paradigm, comprises a two-stage process: feature extraction, and corpus analysis.

In feature extraction, the linguistic structure of the text is analyzed to extract relevant features, and a representation of the text is constructed out of occurrence statistics of these features, usually as a numeric vector. Occurrence statistics can be as simple as raw numbers of the features in the text or their relative frequencies or could include combined statistics like tf-idf (Aizawa 2003) or such measures as "lexical gravity" (Gries & Mukherjee 2010).

The corpus analysis stage takes as input a collection of feature-based document representations and computes a representation of the registers being studied, whether as a classification model (in the categorical paradigm) or a set of dimensions of register variation (in the multidimensional paradigm). Here too, a variety of different technical approaches can be taken, giving different results, even using the same feature extraction method.

**2.2.1 Stylistic Text Features**

A simple and effective set of features (and one of the oldest) for stylistic text analysis is the relative frequencies of function words. Different function words frequencies are indicative of different grammatical choices, and yet are not expected to vary greatly with the topic of the text. For this reason, they are often used for stylistic analysis studies and have proven to be quite efficacious for a variety of such tasks, including register analysis (e.g., Biber 1991, Nowson 2006, Argamon & Levitan 2005, Herring & Paolillo 2006). Typical modern studies using function words in English use lists of a few hundred words, including pronouns, prepositions, auxiliary and modal verbs, conjunctions, and determiners. Numbers and interjections are usually included as well since they are essentially independent of topic, though they are not, strictly speaking, function words.

Another approach is to directly estimate relative frequencies of different syntactic constructions; this has been made possible by development of fast and reasonably reliable natural language parsing techniques. A number of studies have found that using syntactic features can often improve results over traditional word-based analysis alone (Stamatatos et al. 2000). Syntactic structures can be computed using full syntactic parsers (Martin & Jurafsky 2000), but such systems are often brittle when dealing with very long and complex sentences or informal and ungrammatical texts. Shallow parsing, by contrast, seeks just to identify occurrences of certain phrase types (such as noun phrases) without extracting full syntactic structures (Hammerton, Osborne, Armstrong, & Daelemans 2002), and so can be less error prone in those situations, although they provide less-detailed information. Some studies have used even simpler approximations to syntactic features to good effect, such as the frequencies of short sequences of parts-of-speech, or combinations of parts-of-speech and words (Glover & Hirst 1996, Tambouratzis, Markantonatou, Hairetakis, Vassiliou, Tambouratzis, & Carayannis 2000, Koppel, Argamon, & Shimoni 2002).

Taxonomies based on Systemic Functional Linguistics (Halliday & Matthiessen 2004) can be built to represent grammatical and semantic distinctions between classes of words, phrases, and syntactic structures at different levels of abstraction. When applied to lexical items, such taxonomies are a generalization of function word features combined with some parts-of-speech (Argamon et al. 2007). For example, (in English) a pronoun may refer either to a discourse Participant (first or second person) or Non-Participant (third person), if a Participant, it may be the Speaker (*I, me*),

Speaker-Plus (*we, us*), or the Addressee (*you*); if a Non-Participant, it may be an Individual (*he, she, it*), or a Collective (*they, them*); and so on. Such categories form hierarchical taxonomies; the relative frequencies of occurrence within a text of the children of a category show the relative preferences within the text for the different options for realizing the parent category's function. Used as classification features, these relative frequencies have been shown to be effective for register analysis and are particularly helpful for relating the linguistic differences between registers to functional differences in their contexts of use (Argamon et al. 2008, Teich et al. 2013).

The frequencies of character n-grams, while lacking in direct linguistic motivation or interpretation, have proven to be useful in capturing stylistic variation in lexical, grammatical, and orthographic preferences for a variety of tasks, without the need for linguistic background knowledge (making application to different languages trivial). Relative frequencies of character n-grams have proven surprisingly effective for register and genre classification (Kanaris & Stamatatos 2007, Amasyalı & Diri 2006) as well as other style classification tasks such as authorship attribution (Kjell 1995, Kešelj, Peng, Cercone, & Thomas 2003, Stamatatos 2008), document similarity (Damashek 1995), and L1 identification (Koppel et al. 2005). Since character n-grams also capture aspects of document content, however, they must be used with caution and careful experimental control to ensure that the models that are constructed represent register variation and not accidentally correlated topics instead.

**2.2.2 Methods of Register Classification**

For analyzing register in categorical terms, text classification techniques (Sebastiani 2002), usually based on machine learning, are used. The idea is straightforward: training texts are represented as numerical vectors, labeled by their register categories, and machine learning methods are used to find a function that distinguishes between the categories that minimizes some loss function over the training set. Different algorithms will produce different results, with greater or lesser ability to generalize accurately to new data (not in the training set).

A great variety of machine learning algorithms for classification have been applied to stylistic text classification at one time or another, with varying degrees of success. Among the most straightforward are k-nearest neighbor (Kjell et al 1995, Hoorn et al. 1999, Zhao & Zobel 2005) which categorizes according to the label(s) of the nearest document(s) in the training set, and Naive Bayes (Kjell 1994 Hoorn et al. 1999, Peng et al 2004) which chooses the category label with the highest probability, assuming that features occur conditionally independently of each other given the text's category.

Excellent results for register and genre classification have been achieved using diverse machine learning algorithms including support vector machines (De Vel et al. 2001, Diederich et al. 2003, Koppel & Schler 2003, Abbasi & Chen 2005, Koppel et al. 2005, Zheng et al 2006, Sharoff et al. 2010) and neural networks (Matthews & Merriam 1993, Merriam & Matthews 1994, Kjell 1994, Lowe & Matthews 1995, Tweedie et al. 1996, Hoorn 1999, Waugh et al. 2000). Rule-based learning (Holmes & Forsyth 1995, Holmes 1998, Argamon et al. 1998, Koppel & Schler 2003, Abbasi & Chen 2005, Zheng et al. 2006) and Bayesian regression (Genkin et al. 2006, Madigan et al. 2006) have also been applied with some success.

As noted above, the choice of feature set is more important to successful register classification than the specific learning algorithm used. In general, the learning methods that have been found to work best are those that (a) aggregate the impact of many different features together (rather than selecting a small number of "important" features), and (b) are resistant to overfitting, being overly responsive to fine detailed distinctions in the training set (which can lead to reduced performance on new data). These include support vector machines, lasso and Bayesian regression, and some neural network methods.

Register classification is evaluated in the same way as general text classification methods (Sebastiani 2002). A corpus of documents is labeled manually for register categories, and inter-rater reliability is measured. Provided good reliability for labeling is achieved, the corpus is used for training and testing the model, ensuring that testing is done on texts not used for training. Cross-validation (Kohavi 1995) is typically used, in which the full annotated data is divided randomly into several (usually 10) equal-sized "folds" subsets, then training the model on all but one fold and testing on the remaining fold. Results from all the folds are averaged to give an overall effectiveness measure. Effectiveness is measured by classification accuracy as well as average precision and recall over the different register categories (Goutte & Gaussier 2005). These other measures are needed since overall accuracy can be misleading if one category predominates in the data. In such a case, high overall accuracy can be achieved just by classifying all examples by the label of the majority class (i.e., smaller categories will not be accurately recognized at all). Hence, proper evaluation requires considering the precision and recall of each category on its own.

Such "class imbalance" also has implications for classifier training – since most classification algorithms implicitly attempt to optimize for overall accuracy, an imbalanced training set will lead to a model with suboptimal performance. The usual ways to deal with this problem are *undersampling*, randomly keeping only a fraction of examples in the majority category to achieve balanced category sized in the training, or *oversampling*, randomly choosing examples from minority categories to be duplicated in the training set, and hence increase the cost of errors on those categories.

**2.2.3 Methods of Multidimensional Analysis**

In multidimensional analysis, the goal is to find the "natural" dimensions of variation among core grammatical features of the language. Principal Components Analysis (Jolliffe 2011) or Factor Analysis (Loehlin 1998) is used to compute the sets of linguistic features that most frequently co-occur in a corpus. These are called the dimensions of variation for the corpus. Numeric weights are computed for features in each dimension, enabling computation of a score for any text in a given dimension. Analysis of which features covary in each dimension and the relationships between the dimensional scores for different texts or registers enables a linguistic interpretation of how aspects of register variation are represented by the different dimensions.

The earliest work on this approach (Biber 1989) used factor analysis to compute a set of dimensions based on a diverse corpus of English documents, constructed to represent a wide variety of spoken and written registers of British English (mainly from the LOB and LL Corpora, cf. Johansson, Leech, & Goodluck 1978, Svartvik & Quirk 1980). The work used 67 linguistic features in 16 categories including lexical, morphological, and syntactic features. Automatic syntactic analysis and tagging was performed to identify occurrences of each feature, then its frequency per 1,000 words was recorded for each text in the corpus, so that each text was represented by a 67-element numeric vector. Factor analysis was then applied to the corpus to extract the main dimensions of variation. Each factor assigns a weight between -1 and 1 to every linguistic feature – if the feature has a positive weight, it is indicative of a positive value for the factor, if negative, for a negative value. In this way, the factor defines a spectrum for a co-varying set of linguistic features; the value of the factor for a given text represents where that text is on a particular functional linguistic dimension. By examining the commonalities of the features with highly positive and highly negative weights for a factor, we can form an interpretation of what the factor means.

Using this methodology, Biber (1988) identified seven factors of variation related to register, interpreted as:

1. Informational vs. Involved production
2. Narrative vs. Non-Narrative concerns
3. Explicit vs. Situation-Dependent reference
4. Overt expression of persuasion
5. Abstract vs. Non-Abstract information
6. Online informational elaboration
7. Academic hedging

The generality of the result has been supported by the fact that similar studies on other corpora give substantially the same factors (Biber 2003, 2004, Xiao 2009, Clarke & Grieve 2017), though factors differ somewhat in saliency across different corpora, depending on the exact mix of registers and genres present. For example, in Clarke and Grieve's study on Twitter, one dimension primarily reflected tweet length, in that the short length of tweets means that noticeably more varied lexico-grammar appears in longer tweets.

Application of multidimensional analysis to new corpora can either be done by performing a new factor analysis and considering the relationship of the extracted factors to the original factors from Biber (1988), or by using the original factors and loadings to analyze the new corpus directly. Factor scores for documents in different registers or genres of texts give insight into their linguistic and functional content, showing the key distinguishing characteristics of each register.

Conrad's (1996) study of academic biology texts using multidimensional analysis showed that differences between academic and popular nonfiction texts can be quantified in meaningful ways and that important differences within the academic genre can be seen, specifically between textbooks and research articles. Conrad argues that these findings can be used to improve how we teach students to read and understand scientific texts by clarifying the different rhetorical modes that they use. Atkinson (1992) studied the evolution of scientific prose style by (in part) examining how dimensional scores change over time in analysis of the *Edinburgh Medical Journal* (started in 1733). He found clear change over time from somewhat involved style to the more modern informational style and from largely narrative concerns to mostly non-narrative concerns in the most recent articles. The study shows a gradual change of rhetoric and style in scientific communication, arguing that such change is more evolutionary than due to sudden paradigm shifts.

As noted, application of Biber's original method to different sorts of texts and different feature types largely corroborates the original set of dimensions, with some interesting variations at times.

Grieve et al. (2010) used these methods to study linguistic variation in blogs, corroborating Biber (1988) by finding some of the same factors, as well as finding other factors specific to their blog corpus, such as Addressee Focus (whether the blog addresses the reader directly or not) and Thematic Variation (whether blog posts are thematically unitary or address diverse topics).

In another vein, Louwerse and Graesser (2004) used the same methodology of factor analysis but over cohesion features of texts, rather than the lexical and syntactic features used in Biber's original study. Results corroborated the original factors found by Biber, despite the use different features that measured local and global cohesion of the text (Louwerse & Graesser, 2004).

A recent application of the multidimensional paradigm, by Pavalanathan et al. (2017), finds dimensions of linguistic variation related to how interpersonal stances are taken in texts. Their method first generates a lexicon of stance markers, starting from a curated set of seed terms, and then applied multidimensional analysis to frequencies of those markers in a corpus of 530 million Reddit comments. The resulting dimensions include the involved/informational language dimension from Biber's original studies, as well as others more focused on stancetaking such as narrative/dialogue-orientation, standard/non-standard variation, and positive/negative affect.

## 3. Computational Register Synthesis

### 3.1 Aims of Research in Computational Register Synthesis

Computational treatment of register has also been important in text generation, the problem of generating natural text from a structured semantic representation of information to be expressed (Reiter & Dale 2000, Gatt & Krahmer 2018). Text generation systems have been used in a variety of applications, from journalism to presenting healthcare information to intelligent tutoring to multimedia presentation and gaming (Di Eugenio & Green 2010). An important question in text generation for decades has been determining how to generate text that both represents the intended meaning and is also appropriate to a given situation of use. This requires the ability to generate text in different registers for the same propositional meaning (however represented). Two main approaches to this problem have been articulated (cf. Reiter & Williams 2010). The earliest approaches (e.g., Hovy 1988) to generating text in different styles formulated the problem as giving the user control over explicit stylistic features of the text to be generated and applying appropriate constraints to choices made between possible lexical items, syntactic structures, sentence complexities, punctuation, etc. during the text generation process.

A later development was corpus-based methods, that sought to generate texts matching the style of a given body of training texts (a reference corpus). The idea is to run a text generation system in such a way that statistics of linguistic features in the output text will be similar to those of the reference corpus (see, e.g., Langkilde-Geary 2002, Paiva & Evans 2005). This framework allows different styles to be generated based on examples, without explicit determination of the necessary generation parameters, which can be difficult to do well.

### 3.2 Methods of Register Synthesis Research

To synthesize a text in a particular register, choices must be made in how a text is constructed, while still realizing the intended meaning. These choices are inherently integrated within the text generation system, and so we must first discuss how text generation systems work overall. Reiter and Dale (2000) break down text generation into three basic tasks: (a) document planning, determining the overall structure and content of a text, (b) microplanning, determining how the needed information is going to be expressed, in linguistic terms, in sentences and clauses, and (c) realization, generating an actual text conforming to these determinations. Generating text in a particular register amounts to placing constraints (both requirements and preferences) on decisions made at the document planning and microplanning stages.

While some aspects of overall textual organization related to register may be handled at the document planning stage, most aspects of style in extant text generation systems are handled (when they are) during the microplanning stage. These aspects include (Reiter & Williams 2010): lexical choice (which words to use to express information), information aggregation (how much to express in each sentence), information ordering (what order to present information), syntactic choice (what syntactic structures to use), and punctuation choice (full stops vs. exclamation marks, or comma-separated lists vs. numbered lists). When generating text, constraints on these aspects of text construction can be applied to influence the style of the text to be generated. For example, text in a formal register may be generated by constraining lexical choice to prefer formal and Latinate words, information to be more aggregated in fewer (longer) sentences, and syntax to prefer more complex constructions.

The importance of register-based constraints on text generation was realized early in the development of the Penman system (Hovy, Lavid, Maier, Mittal and Paris, 1992) which incorporated a notion of text type, and in the related KOMET system (Bateman, Maier, Teich and Wanner, 1991) which used the notion of a communicative situation. In both cases, and in much subsequent work, the SFL theoretical framework (Halliday & Hasan 1989, Martin 1992) was used to describe register as comprising field (the events, states, and participants in the communicative situation), tenor (the roles and relative statuses of participants), and mode (the kind of communication channel). Text generation systems relate these parameters to constraints on the various linguistic variables described above.

In a text generation system, control over stylistic choices can either be made *explicitly*, by specific control of constraints on document and microplanning, or *implicitly,* by specifying that the system should generate text in a style similar to that of a particular document or corpus. We will discuss each methodology in turn.

In the explicit control scenario, a user explicitly sets the values of a set of parameters that constrain style-related choices in text generation, generally at the microplanning stage. For example, in the Iconoclast system (Power, Scott, & Bouayad-Agha 2003) for generating medical information summaries, the user can set preferred values for:

- Paragraph length,
- Sentence length,
- Frequency of connectives,
- Frequency of passive voice constructions,
- Frequency of pronoun usage,
- Frequency of semicolon usage,

- Frequency of comma usage,
- Level of technical terms allowed, and
- Level of vertical (bulleted) lists allowed.

The system also had aggregated preset values giving higher-level style profiles such as "broadsheet" and "tabloid" that could be selected.

Each of these parameter settings introduces a constraint into the microplanning process, which collectively are expressed as a preference function that scores different microplanning choices accordingly. This will cause the system to make choices that result in longer/shorter paragraphs or sentences, aggregate/separate information to get more/fewer semicolons and commas, and so forth.

In implicit style control, the system is given a document or corpus whose style it is to imitate. One approach to corpus-based style control is manual analysis of the corpus to extract constraints and rules that can be applied during text generation (Hovy 1990, McKeown, Kukich, & Shaw 1994, Reiter & Williams 2010). A difficulty with this approach is that developing such rules is very labor intensive, and the results tend to be of only limited generality.

An approach potentially more useful and generalizable is to measure corpus statistics and use machine learning to generate constraints and rules for text generation in a matching style. Kan and McKeown (2002) describe a text summarization system that learns content planning rules from an annotated corpus for generating summaries matching the style of the corpus. The system uses rule learning (Cohen 1995) to find content-planning rules that match well the lexical statistics (including bigrams and collocations) of the target corpus. A similar approach was taken in SumTime-Mousam (Reiter, Sripada, & Hunter 2005), which created microplanning rules for generating weather forecasts based on machine learning over a corpus of human-authored forecasts. These rules were, however, filtered by domain experts before they were used in the system. Belz (2005) describes a version of the system that used automatically learned rules without any human input – the results were not as good as those of SumTime-Mousam, but were still judged to be quite readable and understandable.

The corpus-based approach, by contrast, seeks to generate texts matching the style of a given reference corpus without explicit manual determination of generation parameters. A simple, but time-consuming, method is to search systematically through different parameter settings for the text generation system, comparing the various texts generated with different settings to the reference corpus, and picking the settings that produces a text with the most similar feature statistics (cf. Langkilde-Geary 2002). A more long-term efficient methods is described by Paiva & Evans (2005), whose system first analyze the reference corpus to infer the parameter settings most likely to produce a style consistent with it, and then use those predetermined settings to generate new texts. The first step is done by generating a variety of texts for multiple parameter settings and analyzing the correlations between choices taken in generation and similarity scores between generated texts and the reference corpus. The result is a score for each possible choice in generation for how likely it is to generate texts that match the reference corpus; the most likely choices can then be used to generate texts statistically matching the reference style.

Related to this corpus-based approach for generating text in a particular register are techniques for directly "translating" texts from one register or style to another. The general goal is to take a text in one register (or style) and generate text expressing the same content, but in a different register. One of the earliest such works in Xu, Ritter, Dolan, Grishman, & Cherry (2012) work on paraphrasing modern texts in Shakespeare's style, which treated the problem as one of machine translation (MT), applying a standard phrase-based statistical MT method (Och & Ney 2003, Koehn et al. 2007). The task was formulated by taking the original versions of the plays and learning a translation model between them and translations of them into Modern English. The learning process found correspondences between words and phrases, such as "thou" → "you", as well as probabilities of changing word order. They applied the model to a play not used in training (*Romeo and Juliet*) and compared results to human translations.

Recently, research has grown on this "style translation" task using neural network methods, addressing it as a variant of the "style transfer" problem for images (Gatys et al. 2016), using sequence-to-sequence networks, whose input and output are each a sequence of tokens. Variables denoting the desired style are added as parameters either as special tokens in the input text (Sennrich et al. 2016) or as extra inputs to the neural network (Ficler & Goldberg 2017). Typically, these approaches use Long Short-Term Memory (LSTM) networks (Hochreiter & Schmidhuber 1997). An LSTM network is given each token in a text (usually a sentence) as a vector-encoded input one at a time and computes both an update to a set of memory units and a vector-encoded output token as a function of the input and the current memory state. It thus directly translates one sequence of tokens into another sequence of tokens. Training is done by comparing network outputs to gold-standard training data (what sequence should result) and adjusting weights in the network to make outputs more similar to the training for the given inputs.

For style translation, an LSTM network is trained on input/output pairs of sentences in two different styles (say, Shakespeare and a modern translation), with addition inputs encoding which pair of styles is being trained on (so the network can automatically generate the correct style after it is trained). For generation, these style parameters will influence the probability distribution of lexical items generated and the length/complexity of sentences generated by the network, so it will generate sentences in a particular style. Improvements have been made by applying adversarial network training (Goodfellow et al. 2014), in which the style translating network tries during training to fool a second network that determines whether texts are in the target style or not. This training method has been applied recently to improve accuracy at style translation in this paradigm (Prabhumoye, Tsvetkov, & Salakhutdinov 2018, Fu et al. 2018).

# 4. Case Studies

To make the discussion of different research goals, methods, and results more concrete, in this section I will describe three case studies, one on register classification, one on multidimensional analysis, and one on register synthesis.

## *4.1 Classification analysis: Multidisciplinary scientific texts*

In an excellent exemplar of classification analysis of register characteristics, Teich and Fankhauser (2009) consider the question of how clusters of linguistic characteristics emerge in new registers as multidisciplinary scientific fields coalesce. To do this, they analyze texts from the DaSciTex corpus which comprises journal articles from nine scientific fields, including interdisciplinary ('mixed') fields. They are:

>  **A** computer science
>  **B** 'mixed' disciplines:
>  >  **B1:** computational linguistics
>  >  **B2:** bioinformatics
>  >  **B3:** computer aided design/construction in mechanical engineering
>  >  **B4:** microelectronics/VLSI
>  
>  **C** 'pure' disciplines
>  >  **C1**: linguistics
>  >  **C2**: biology
>  >  **C3**: mechanical engineering
>  >  **C4**: electrical engineering

The research investigated how the registers of the 'mixed' (B) disciplines relate to those of their respective core 'pure' disciplines (C) and that of computer science research (A) which they draw from. This study provides a good case study of how classification methods are used for register analysis, and the kinds of broader conclusions that can be drawn from such analysis.

The study comprises several analyses to determine the character of the registers being examined. First, the overall characteristics of the scientific register of the DaSciTex corpus of scientific articles is analyzed by comparison with the Freiburg–LOB Corpus of British English (FLOB). The authors focused on linguistic features likely to indicate theoretically predicted characteristics of scientific text: abstractness, technicality, and informational density. These included the fractions of words that are common nouns, lexical verbs, and adverbs; the type-token ratio to indicate use of specialized terminology; and lexical density as a proxy for informational density. Discriminability measures such as information gain (Quinlan 2014, Raileanu & Stoffel 2004) show which features individually most distinguish the two categories (scientific and general text); as expected higher abstractness, information density, and terminological diversity (lower type-token ratio) are characteristics of scientific text. The discriminating power of these features together was explored by using them as input into support vector machine classification and k-means clustering; these gave high classification accuracy indicating strong differences between the document groups. Additionally, these features do not strongly distinguish between different subcategories of the DaSciTex corpus, indicating that all these scientific registers are similar in these respects.

The authors analyze the register characteristics of the different scientific disciplines within the DaSciTex corpus by learning linear classifiers (using support vector machines) to distinguish articles from the different disciplines by occurrence frequencies of distinctive (measured by information gain) nouns and lexical verbs, respectively. Classification accuracy is high, 96%, and 87%, respectively. Examination of the confusion matrices, which show how often documents in one category were classified as each of the others, enables measuring the linguistic similarity of different document categories – categories that are more often confused with each other are more similar. Results here showed that the language of the mixed disciplines are each most similar to their respective pure disciplines (B1 and C1, B2 and C2, etc.), and somewhat to computer science (A), and also electrical engineering (C4) is similar to computer science overall.

A more fine-grained analysis of the computer science (A), computational linguistics (B1), and linguistics (C1) subcorpora was done by learning linear classifiers using as features the different verbs that appear with "we" as a subject in the articles. These show what activities the authors are portrayed as engaging in by the articles. The most significant such features for computer science were formal activities (e.g., "prove" and "define"); for computational linguistics were experimental (e.g., "collect" and "examine"); and for linguistics were communicative (e.g., "argue" and "read") and cognitive (e.g., "see" and "feel"). This gives some insight into how scientists in the different disciplines understand the key activities of their disciplines, and how they construct themselves as acting, in their research reports.

Classification analysis thus can give multiple views into the register variations under study, by a combination of univariate and multivariate results, and exploration of classification accuracy with misclassification rates for different feature sets and subsets of documents. As Teich and Fankhauser comment:

> In addition to ranking features by their individual discriminatory power, we can explore their *collective* contribution to register discrimination (multivariate analysis). Also, having available information about misclassifications in the form of the confusion matrix, we can investigate the context of typical features/terms in correctly classified and in misclassified texts, analyzing differences and commonalities between registers at class level as well as at instance level. (Teich & Fankhauser 2009: 245; italics in original)

### *4.2 Multidimensional analysis: Abusive language*

Multidimensional analysis was used recently to good effect by Clarke and Grieve (2017) in a study of racist and sexist language on Twitter. They study a corpus of 16,914 Tweets from the Waseem and Hovy (2016) Twitter corpus that have been manually coded as racist (1972 tweets), sexist (3383 tweets), or neither (11,559 tweets), with the goal of understanding register variation between abusive and non-abusive tweets and the similarities and differences between racist and sexist language on Twitter.

Due to the short length of tweets (typically under 30 words), standard multidimensional analysis (MDA) cannot be applied, since it relies on measuring and comparing feature frequencies within each text, and any given feature might appear once in a given tweet, if at all. The authors therefore use a variation called multiple correspondence analysis (MCA) (Husson, Lê, & Pags 2010) which reduces categorical data to underlying dimensions in a similar way to how MDA reduces quantitative data. Linguistic features in the tweets are therefore treated as categorical variables (occurrence vs. non-occurrence) and MCA used to extract underlying dimensions of variation. The analysis used 81 linguistic features that occurred in at least 1% of the tweets, including "tense and aspect markers, place and time adverbials, personal pronouns, questions, nominal forms, passives, subordination, complementation, adjectives and adverbs, modals, specialized verb classes, coordination, negation and other lexical classes, such as amplifiers, down- toners and conjunctions." (Clarke & Grieve 2017) Additionally, they included medium- and genre-specific features such as "hashtags, URLs, capitalization, imperatives, comparatives, and superlatives."

Clarke and Grieve extract four dimensions from the corpus using MCA, however the first dimension is strongly correlated with tweet length. As one might expect, since tweet length is small on average with a high standard deviation, a key dimension of variation will be text length. Hence, they remove that dimension from consideration in further analysis. The other three dimensions are interpreted as:

1. Interactivity: Variation between tweets directly interacting with others and those simply asserting information. Features of interactive tweets include question marks, question and clause-initial DO, WH-words, and first and second person pronouns; while those of non-interactive (informational) tweets include existential *there, be* as a main verb, numbers and attributive adjectives, quantifiers, prepositions, and proper nouns, as well as other features of informational text such as nominalizations and contrastive conjunctions.
2. Antagonism: Ranging from aggressive and antagonistic tweets to more those of more conciliatory style. Features of antagonistic tweets include those indicating direct confrontation and emotion, such as question and initial DO, clause-initial verbs, question marks, and second person pronouns, as well as nominalizations, possessive pronouns, emoticons, and exclamation marks. Conciliatory features included lack of second person pronouns, especially with first and third person pronouns in subject and object position, progressive verbs, and perception verbs.
3. Attitudinality: Variation in whether a tweet explicitly expresses and attitude, or frames itself as factually propositional. Features indicating attitudinality include comparatives, predicative adjectives, first person pronouns, existential *there*, and paucity of nouns; indicating factuality include public verbs mark indirect or reported speech, marked aspect (perfect or progressive), passive voice, URLs, and numbers.

These three dimensions define a space of register variation within which the abusive (racist and sexist) tweets exist. Further analysis of the distributions of dimension scores for racist and sexist tweets shows some interesting differences. Data visualization and the Wilcoxon signed-ranked test showed sexist tweets to be more interactive and attitudinal than racist tweets, on average. The authors interpret these findings in light of Herring et al.'s (1995) study of silencing strategies in sexist discourse as the sexist tweets perhaps aiming to take control of the Twitter discussion and silence female voices, while racist tweets serve more to spread racist ideology by storytelling and persuasion, in the modes described by van Dijk (1993).

This study shows how multidimensional analysis applied to multiple related document categories can elucidate the linguistic structure of register variation and contribute to understanding the potential functions of such variation.

### *4.3 Text generation: Customized medical information*

As a case study for text generation in variable register, consider WebbeDoc (DiMarco & Foster 1997), a system developed to produce customized information about the HealthDoc medical text generation project. The idea is to present the same information about the HealthDoc project in different ways depending on reader characteristics such as professional role (physician, computational linguist, funder, layperson), age, and reading level. The work does not explicitly deal with the notion of register but rather with aspects of textual style that in the aggregate amount to different registers for different audiences and purposes.

WebbeDoc works from an internal structured representation of a complete "master document" about the project, which can be realized in different kinds of actual texts. This internal master document (MD) contains elements which are abstract sentence specifications (written in a structured Text Specification Language, or TSL) that express elemental propositions. These are connected within a structure that represents ordering constraints, rhetorical relations, coreference links, and formatting information (such as use of vertical lists or illustrations). Each element is also annotated with user model constraints, labeling the user categories (e.g., "layperson") or textual styles (e.g., "high technical level" or "informal") for which the element should be selected. The master document as a whole is structured into topics and subtopics, each of which can be divided into "version sets" which contain the different ways to tailor the information in the topic for different audiences (per the selection annotations). In text generation, document planning will choose between versions of each subtopic, appropriate to the given parameter settings. Preferred ordering of subtopics may also vary with intended audience and style.

Document planning seeks to construct a coherent text expressing the information in the master document relevant to a given audience in an appropriate style. The system chooses to use those specific MD elements whose annotations are consistent with the parameters specified, and then seeks to satisfy consistent ordering preferences, and general rhetorical and coreference constraints to ensure completeness and coherence of the information to be presented. Constraint satisfaction is used to find such a coherent information structure to present (Marcu 1997). Aggregation of matching entities is applied to simplify syntactic structures and determine appropriate use of coreferential expressions, and then text is generated from the internal representation. A variety of repair strategies is applied to ensure the consistency and coherence of the resulting document.

Unfortunately, as with much other work in customizable text generation, no user studies were performed on WebbeDoc. However, examples show the plausibility of generated texts for different audiences.

# 5. A Program for Future Research

Computational research on register has the potential to both benefit from and contribute to more elaborated and precise theoretical and empirical analyses of register in general. As we have discussed, notions of register and genre, especially as used in computational language research, are often vague and imprecise, relying on intuitive categories of "text types" or small sets of register categories that are different enough to be easily distinguished, such as "spoken" vs. "written", or "news articles" vs. "research papers". The most consistent and clear approach to date has been Biber's multidimensional approach, with a clear empirical basis and elaborated dimensions based on well-accepted linguistic features. However, there is as yet no general model relating these dimensions directly to aspects of the communicative situation (as in the SFL theory of register), to accepted register categories, in the categorical approach, or to the choices made by a text generation system that realizes a text in a given register. It is here that computational research into register variation and connected phenomena can make a fundamental contribution to the understanding of register and the connection between the communicative situation and linguistic style and form.

In broad outline, the research program I suggest is to use and combine existing and new computational register analysis and synthesis methods to elaborate and test detailed and empirically based models of register-based linguistic variation. (The program is closely related to Matthiessen's [2015] registerial cartography; see further below.) If we consider the parameters and constraints that determine the form of a text (see Figure 2), they include many disparate elements, many of which feed into traditional notions of register, such as the medium of communication, the purpose of the text and intended audience, ideological roles/relationships among the author, editor(s), and audience, and the intertextual context of the text. The sort of precise and elaborated model envisioned here would describe causal relationships between parameterizations of these elements and the linguistic form of the text, likely mediated by intermediate variables such as Biber's dimensions and register/genre categories at different levels of specificity. Such models, perhaps in the form of hierarchical probabilistic models (Koller & Friedman 2009) or structured causal models (Pearl 2009), would enable precise empirical predictions from models of how the social and communicative context influences the construction of different texts and gives rise to different registers.

An immediate research goal is to put the notion of register categories, or text types, on a firmer empirical footing. Linking work on register categorization with multidimensional analysis would be an excellent step – exploring how register categories might be viewed as regions within a multidimensional "style space", and their relationships measured and compared. Rigorous work comparing the dimensions that emerge from analysis of different corpora would be needed, so that a standardized dimensional representation could be determined, or at least that regularities in how to understand any variation in the extracted dimensions. (It seems likely that one underlying set of dimensions can be derived, with variation attributable to corpus composition, but this will require some work to prove.) Comparison between models in different languages and cultures will be necessary, with the goal of finding typological universals; most work has been done in English, though work in other languages is consistent with multidimensional analyses in English (Biber 1995, Berber Sardinha 2017). This work will give a clearer basis for the descriptive analysis of register, enabling improved large-scale empirical work in the area.

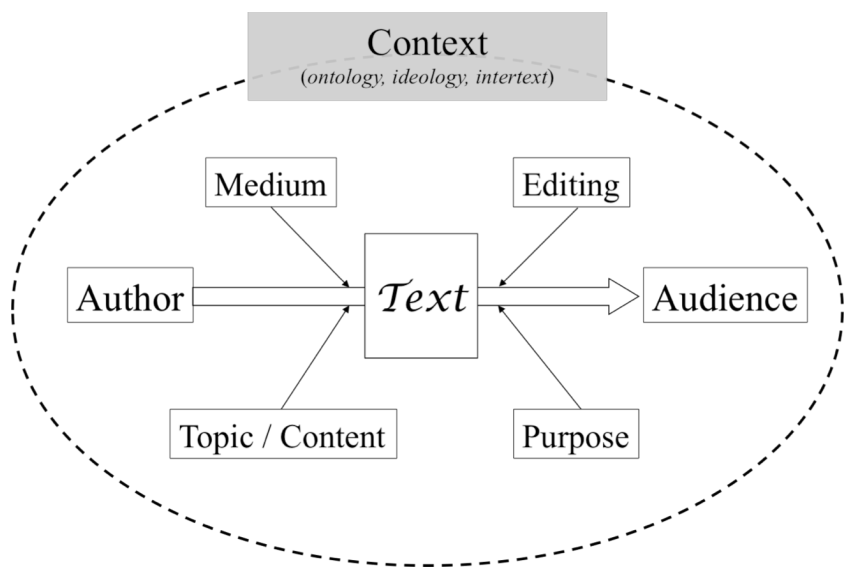

Figure 2. High-level view of parameters and constraints on textual form and style. Adapted from Argamon & Koppel (2010).

A second key focus for connecting disparate research in the area is integrating register analysis and synthesis; while both active research areas use similar theoretical concepts of register and linguistic variation, there is no cross-over between the research, and so no clarity as to the relationships between register dimensions, occurrence of linguistic features, and planning decisions during text generation. Register classification systems can be used to evaluate the results of text generation systems, and correlated with user perception studies, to better validate schemes for categorization and measurement of register variation. Comparison of Biber's multidimensional model and the internal representations learned by style transfer neural networks may also help elaborate the ways in which linguistic representation space can usefully be structured to represent situational variation.

As clearer and more integrated representation schemes are developed for register varieties and relationships between them, register can be more easily addressed properly within work on natural language processing (NLP) and computational linguistics. Much work in these fields, as noted above, does not properly account for register, or does so superficially by controlling for some prelabeled "text types" or the like. Given broader articulated models for register, and tools to implement register classification and profiling, more sophisticated methods for register-aware evaluation of NLP tools can be developed, and better experimental controls for register can be applied.

Furthermore, these same methodological and computational tools will enable qualitatively greater progress in the large-scale descriptive task termed by Matthiessen (2015) as registerial cartography – mapping the functional language varieties in a language/culture and the relationships between them. More sophisticated computational analysis tools will enable larger scale studies than have been heretofore possible, and unified representational schemes will allow comparison and integration of the results of different studies on different corpora. Historical and typological studies will also allow for the descriptive study of register evolution at the large scale, expanding greatly the scope of such studies as Atkinson's (1992) of the development of the modern scientific register and Gries and Mukherjee's (2010) work on the evolution of World Englishes. As Matthiessen notes, development of the theory of register beyond a certain point requires the detailed description and mapping of many extant register varieties and the relationships among them, to ground the theory. (See also Matthiessen and Teruya [2015] for a good example.) One grand challenge for computational research on register, therefore, will be to develop the representations and tools necessary for this task at a large scale.

The other grand challenge will be the development of the methodological and computational tools necessary for empirical verification of theories of register. This would seem to require the ability to specify clear, articulated models of the causal relationships between situational parameters (social roles and relationships, communication medium, etc.), linguistic features, and intermediate representational levels, and then to test these models empirically, either by analysis of large corpora, or by generating texts according to specification and rigorously measuring human responses to the texts.

# References

Argamon, S., Koppel, M., Fine, J., & Shimoni, A. R. (2003). Gender, genre, and writing style in formal written texts. *Text, 23*(3), 321-346.

Argamon, S., Whitelaw, C., Chase, P., Hota, S. R., Garg, N., & Levitan, S. (2007). Stylistic text classification using functional lexical features. *Journal of the American Society for Information Science and Technology, 58*(6), 802-822.

Atkinson, D. (1992). The evolution of medical research writing from 1735 to 1985: The case of the Edinburgh Medical Journal. *Applied Linguistics, 13*(4), 337-374.

Bateman, J. A., Maier, E. A., Teich, E., & Wanner, L. (1991). Towards an architecture for situated text generation. In *Proceedings of the ICCICL* (pp. 289-302).

Belz, A. (2005). Statistical generation: Three methods compared and evaluated. In *Proceedings of ENLG-2005* (pp. 15–23).

Berber Sardinha, T. (2017). Text types in Brazilian Portuguese: a multidimensional perspective. *Corpora, 12*(3), 483-515.

Biber, D. (1991). *Variation across speech and writing*. Cambridge University Press.

Biber, D. (1995). *Dimensions of register variation: A cross-linguistic comparison*. Cambridge University Press.

Biber, D. (2003). Variation among university spoken and written registers: A new multi-dimensional analysis. *Language and Computers, 46*, 47-70.

Biber, D. (2004). Conversation text types: A multi-dimensional analysis. In *Le poids des mots: Proc. of the 7th International Conference on the Statistical Analysis of Textual Data, Louvain: Presses universitaires de Louvain* (pp. 15-34).

Biber, D. & Barbieri, F., 2007. Lexical bundles in university spoken and written registers. *English for specific purposes, 26*(3), pp.263-286.

Biber, D., & Conrad, S. (2001). Register variation: A corpus approach. In D. Schiffrin, D. Tannen, & H. E. Hamilton (Eds.) *The handbook of discourse analysis* (pp. 175-196). Malden, MA & Oxford: Blackwell.

Biber, D., & Conrad, S. (2009). *Register, genre, and style*. Cambridge University Press.

Biber, D., & Finegan, E. (2001). Diachronic relations among speech-based and written registers in English. In S. Conrad & D. Biber (Eds.) *Variation in English: Multi-dimensional studies* (pp. 66-83). Harlow, England: Pearson Education.

Biber, D., 1989. A typology of English texts. *Linguistics, 27*(1), pp.3-44.
A version of this article is to appear in *Register Studies,* 2019.                                23

Johansson, S., Leech, G. N., & Goodluck, H. (1978). *Manual of information to accompany the Lancaster-Oslo/Bergen Corpus of British English, for use with digital computer*. Department of English, University of Oslo.

Jolliffe, I. (2011). Principal component analysis. In *International encyclopedia of statistical science* (pp. 1094-1096). Berlin, Heidelberg: Springer.

Kakkonen, T., & Sutinen, E. (2008). Coverage-based Evaluation of Parser Generalizability. In *Proceedings of the Third International Joint Conference on Natural Language Processing: Volume-II*.

Kan, M.Y. and McKeown, K.R., 2002. Corpus-trained text generation for summarization. In *Proceedings of the International Natural Language Generation Conference* (pp. 1-8).

Kanaris, I., & Stamatatos, E. (2007). Webpage genre identification using variable-length character n-grams. In *Proceedings of the 19th IEEE International Conference on Tools with Artificial Intelligence* (p. 3-10), Washington, DC.

Karlgren, J. (1999). Stylistic experiments in information retrieval. In T. Strzalkowski (Ed.) *Natural Language Information Retrieval* (pp. 147-166). Dordrecht: Springer.

Kešelj, V., Peng, F., Cercone, N., & Thomas, C. (2003). N-gram-based author profiles for authorship attribution. In *Proceedings of the Conference Pacific Association for Computational Linguistics, PACLING, 3*, 255-264.

Kjell, B. (1994a), Authorship attribution of text samples using neural networks and Bayesian classifiers. In IEEE International Conference on Systems, Man and Cybernetics, San Antonio, TX.

Kjell, B., Woods, W. A., Frieder, O. (1995), Information retrieval using letter tuples with neural network and nearest neighbor classifiers. In IEEE International Conference on Systems, Man and Cybernetics, volume 2, pp. 1222-1225, Vancouver, BC.

Koehn, P., Hoang, H., Birch, A., Callison-Burch, C., Federico, M., Bertoldi, N., Cowan, B., Shen, W., Moran, C., Zens, R., Dyer, C., Bojar, O., Constantin, A., & Herbst, E. (2007). Moses: open source toolkit for statistical machine translation. In *Proceedings of the 45th Annual Meeting of the ACL on Interactive Poster and Demonstration Sessions*.

Kohavi, R. (1995). A study of cross-validation and bootstrap for accuracy estimation and model selection. In C. S. Mellish (Ed.). *Proceedings IJCAI-95, 14*(2), 1137-1145. Montreal, Quebec.

Koller, D., Friedman, N., & Bach, F. (2009). *Probabilistic graphical models: Principles and techniques*. MIT press.

Koppel, M. and Schler, J., 2003, August. Exploiting stylistic idiosyncrasies for authorship attribution. In *Proceedings of IJCAI'03 Workshop on Computational Approaches to Style Analysis and Synthesis* (Vol. 69, pp. 72-80).
A version of this article is to appear in *Register Studies,* 2019.        27